\documentclass[letterpaper, 10 pt, conference]{ieeeconf}  

\IEEEoverridecommandlockouts                              

\title{\LARGE \bf
Efficient Human-Aware Task Allocation\\ for Multi-Robot Systems in Shared Environments
}

\author{Maryam Kazemi Eskeri, Ville Kyrki, Dominik Baumann, and Tomasz Piotr Kucner
\thanks{This work was supported by the Research Council of Finland Flagship programme: Finnish Center for Artificial Intelligence FCAI.
We acknowledge the computational resources provided by the Aalto Science-IT project and CSC, Finnish IT Center for Science.}
\thanks{The authors are with the Department of Electrical Engineering and Automation, Aalto University, 02150 Espoo, Finland (e-mail: firstname.lastname@aalto.fi)
}
}%

\usepackage[table]{xcolor}
\usepackage{cleveref}
\usepackage{booktabs}
\usepackage{multirow}
\usepackage{graphicx}
\usepackage{pgf}
\usepackage{pgfplots} 
\usepackage{siunitx}
\usepackage[acronym]{glossaries}
\usepackage{mathtools}
\usepackage{everypage}
\usepackage{stfloats} 
\usepackage{float}
\usepackage[final]{changes}
\usepackage{adjustbox}

\makeatletter
\newcommand{\acposs}[1]{%
 \expandafter\ifx\csname AC@\AC@prefix#1\endcsname\AC@used
   \acs{#1}'s%
 \else
   \aclu{#1}'s (\acs{#1})%
 \fi
}
\makeatother

\newacronym{method name}{HATA}{Human-Aware Task Allocation}

\usepackage{fancyhdr}
\newcommand{\mytitle}{\textbf{Accepted final version.}
To appear in \textit{Proceedings of the IEEE}.\\
\copyright 2025 IEEE. Personal use of this material is permitted. Permission
from IEEE must be obtained for all other uses, in any current or future
media, including reprinting/republishing this material for advertising or
promotional purposes, creating new collective works, for resale or
redistribution to servers or lists, or reuse of any copyrighted component of
this work in other works.}
\begin{document}

\maketitle
\thispagestyle{empty}
\pagestyle{empty}

\begin{abstract}

\added{Multi Robot Systems are increasingly deployed in applications, such as intralogistics or autonomous delivery, where multiple robots collaborate to complete tasks efficiently. One of the key factors enabling their efficient cooperation is Multi-Robot Task Allocation (MRTA). Algorithms solving this problem optimize task distribution among robots to minimize the overall execution time. In shared environments, apart from the relative distance between the robots and the tasks, the execution time is also significantly impacted by the delay caused by navigating around moving people. However, most existing MRTA approaches are dynamics-agnostic, relying on static maps and neglecting human motion patterns, leading to inefficiencies and delays. In this paper, we introduce \acrfull{method name}. This method leverages Maps of Dynamics (MoDs), spatio-temporal queryable models designed to capture historical human movement patterns, to estimate the impact of humans on the task execution time during deployment. \acrshort{method name} utilizes a stochastic cost function that includes MoDs Experimental results show that integrating MoDs enhances task allocation performance, resulting in reduced mission completion
times by up to $26\%$ compared to the dynamics-agnostic method
and up to $19\%$ compared to the baseline. This work underscores the importance of considering human dynamics in MRTA within shared environments and presents an efficient framework for deploying multi-robot systems in environments populated by humans.}
\deleted{Multirobot systems operating in populated environments need to account for humans to enhance efficiency in task allocation. Currently, the most efficient Multi-Robot Task Allocation approaches rely on static maps without considering patterns of dynamics. This results in suboptimal decisions, such as assigning tasks without considering potential delays, leading to inefficient task execution and longer completion times. In this paper, we propose a method that includes information about dynamics into the task allocation cost function. \acrshort{method name} adopts a stochastic cost function incorporating both path length and the likelihood of human encounters, whilst Bayesian optimization is utilized to estimate optimal weighting parameters for the cost function. To measure the performance improvements, we evaluate our framework with realistic pedestrian motion data through extensive experiments. The results confirm that our human-aware task allocation technique significantly reduces robot waiting time and shortens overall mission duration, achieving a notably higher success rate in comparison with traditional allocation mechanisms, especially for cases where human density is relatively high.}

\end{abstract}

\section{INTRODUCTION}

\added{Many real-world scenarios require multiple actors to cooperate to achieve a common objective effectively. Applications, like search and rescue \cite{drew2021multi}, or intralogistic \cite{bolu2021adaptive},
often involve multiple agents coordinating to complete complex tasks.
Multi-Robots Systems (MRS) provide an effective solution by distributing workload and enabling parallel execution.
Today, MRS are widely applied in human-shared environments, like service robots, 
or pick-up and delivery~\cite{chakraa2023optimization}. However, operating in dynamic environments poses critical challenges as robots need to complete tasks efficiently and safely while navigating human-populated spaces.}
One of the main challenges in MRS is the Multi-Robot Task Allocation (MRTA), which is the problem of deciding which robot should execute which task to optimize overall system performance. Performance is typically assessed by the total completion time~\cite{gerkey2004formal}. When MRS are operating in dynamic environments, the configuration of the space is constantly changing because of semi-static obstacles or humans. These dynamics introduce unpredictability, requiring robots to account for them in their decision-making processes. Therefore, task allocation in dynamic environments represents a challenging scenario for MRTA problem \cite{chakraa2023optimization}.

\added{In recent years, numerous studies have been conducted addressing MRTA \cite{chakraa2023optimization, dai2025heterogeneous}; however, many of these studies fail to account for the impact of dynamic entities (eg. human) moving within the environments. Consequently, these approaches often result in significant delays or even failure for robots to accomplish their goals \cite{surma2021multiple}. In environments where robots must navigate around people, it is crucial to account for scenarios where robots may need to wait or adjust their paths to avoid collisions with people. If robots are allocated to tasks based on a dynamics-agnostic method, like shortest path approaches, robots need to react locally to dynamic entities they sense in the environment. Accounting for dynamics only after observation can lead to inefficiency, highlighting the need for incorporating long-term human motion patterns to improve the efficiency of MRS in environments populated with humans.}


We aim to address this gap by proposing a \gls{method name} method that incorporate human motion patterns into the allocation process. To this end, we leverage Maps of Dynamics (MoDs)~\cite{kucner2023survey}, which are spatial or spatio-temporal queryable models of motion patterns. The main contributions of this paper are: (1) Introducing a stochastic cost function that balances path length and human encounter likelihood along the path, enabling robots to account for the impact of delays and collisions and minimize their occurrence. (2)  Proposing a scalable framework for human-aware task allocation that significantly reduces computational overhead compared to the baseline~\cite{surma2021multiple}. \deleted{Experimental analysis demonstrates that as the number of robots grows, the computational advantage of \acrshort{method name} over the baseline becomes more pronounced.}
 \deleted{ evaluated using experimental analysis. These experiments demonstrate that as the number of robots increases, the gap in computational overhead between the baseline method and \acrshort{method name} increases.}
(3) Evaluating \acrshort{method name} in extensive simulation experiments demonstrating \acrshort{method name}'s competitive success rate, mission, and waiting time against two baselines. \added{Our evaluation framework incorporates real human motion data to simulate human-populated environments and utilize a coordination framework to manage the interaction between robots and robot-pedestrians. To our knowledge, this is the first work to evaluate MRTA in such a comprehensive manner. }

\section{RELATED WORK}
There has been extensive research on addressing MRTA in various scenarios. ~\cite{verma2021multi} have classified methods for solving the MRTA problem into centralized and decentralized.
In centralized methods, a dispatcher with complete knowledge of the entire fleet handles the assignments~\cite{agarwal2019cannot}.
While these methods can generate optimal solutions, they are at risk of a single-point failure. Additionally, as the number of robots and tasks increases, the computational load increases substantially, leading to scalability issues and making the problem intractable~\cite{korsah2013comprehensive}. Decentralized architectures address these challenges by distributing the computational load among multiple agents~\cite{verma2021multi}. Among decentralized approaches, auction-based methods, first  This shift has led to interaction-aware planning methods, though significant challenges remain - particularly the "freezing robot problem" where robots become paralyzed by uncertainty in crowded, dynamic environments., have been widely adopted for the distributed task allocation problem. In auction-based approaches, agents compete for tasks through a bidding process, where agents bid on tasks and allocation is based on maximizing auction revenues~\cite{liu2023asynchronous}. These methods find many applications because of their easy implementation, fast convergence, computational efficiency, and scalability~\cite{otte2020auctions}.
However, most current applications of auction-based methods utilize a simplistic cost function in which the dynamic nature of the environment is not fully captured. These methods often rely on measuring static parameters such as distance~\cite{sarkar2018scalable}, or travel time estimates based on a constant velocity assumption~\cite{de2022decentral},
neglecting the temporal and spatial variations of real-world environments. \deleted{Thus, these limitations result in sub-optimal solutions in scenarios where human-robot interactions are frequent.}

Addressing these limitations requires auction-based methods to incorporate information about dynamics into their allocation. 
\added{Some researchers attempt to address task allocation in environments where agents are subjected to external forces, such as wind or ocean currents.} \deleted{dynamic environments where motion follows structured and predictable patterns.} For example, \cite{bai2018integrated} proposes a method for task allocation in drift fields, where agents operate under continuous environmental forces. Their approach integrates optimal control and a genetic algorithm to efficiently assign tasks while minimizing travel time.\deleted{ While effective for such scenarios, it does not address highly stochastic settings where agent movement is influenced by human behavior rather than physical laws, which highlights the need for approaches specifically designed to handle human-influenced dynamics.}
\added{Such methods focus on the impact of the medium on the travel time in fairly simple environments (i.e., sea, air), yet they are not applicable for scenarios where dynamics are caused by the presence of discreet macroscopic objects such as humans.}

\deleted{in structured environments, this method relies on predictable flow patterns, making it unsuitable for highly stochastic settings where agent movement is influenced by human behavior rather than physical laws.}

\deleted{Recent work by~\citeauthor{surma2021multiple}~\cite{surma2021multiple} tries to address this by integrating human motion patterns into the auction process. While their approach, termed HA-Alloc, improves task allocation efficiency in dynamic environments, the computational overhead is significantly high, which poses scalability issues.}

\added{To the best of our knowledge, \cite{surma2021multiple} is the most recent approach to human-aware task allocation in dynamic environments. Their method, HA-Alloc, integrates human motion patterns into the auction process to improve task allocation efficiency. However, despite its effectiveness, HA-Alloc has a significant computational overhead, which raises scalability concerns. Additionally, the cost function weights are estimated based on simulated travel times, rather than using a multi-robot coordination simulator that accounts for interactions between robots. As a result, the accuracy and generalization of the estimated weights remain uncertain. The evaluation is also limited to small-scale experiments with up to seven robots, leaving its performance in larger fleets unexplored. These limitations highlight the need for a more efficient and scalable solution that comprehensively incorporates information about dynamic entities in the environment.}

We propose a method that leverages Maps of Dynamics (MoDs)~\cite{kucner2023survey}, spatio-temporal models that represent motion patterns over time. MoDs are not limited to the robot's perceptual range and provide a comprehensive view by modeling global motion patterns. This capability supports downstream tasks (eg. task allocation), to make informed decisions by considering long-term trends and potential future states of the environment. By incorporating such \deleted{predictive capabilities} \added{models}, robots can better anticipate potential interactions with moving entities and navigate through dynamic environments, improving their decision-making processes.



MoDs can be categorized based on the types of dynamics they represent.
In broad strokes, we can split all existing objects into static, which do not change their position during the robot's mission, and dynamic, which do change their position. The dynamic objects can be further split into active objects that do not remain in a single state for extended periods and semi-static objects that tend to remain in few conﬁgurations within the robot’s environment. 
Various approaches have been developed to model active entities' behavior, including methods that observe velocities (e.g., CLiFF-map~\cite{kucner2017enabling}), or predict trajectories~\cite{yuan2017review}.

In this work, we focus on modeling dynamic obstacles like pedestrians using a grid-based model capturing the probability of human presence. By incorporating this representation into the bidding process, we enable the task allocation method to anticipate and account for potential interactions with humans in the environment. 
\begin{figure*}[h!]
    \centering
    \includegraphics[trim=10 450 10 10, clip, scale=0.22]{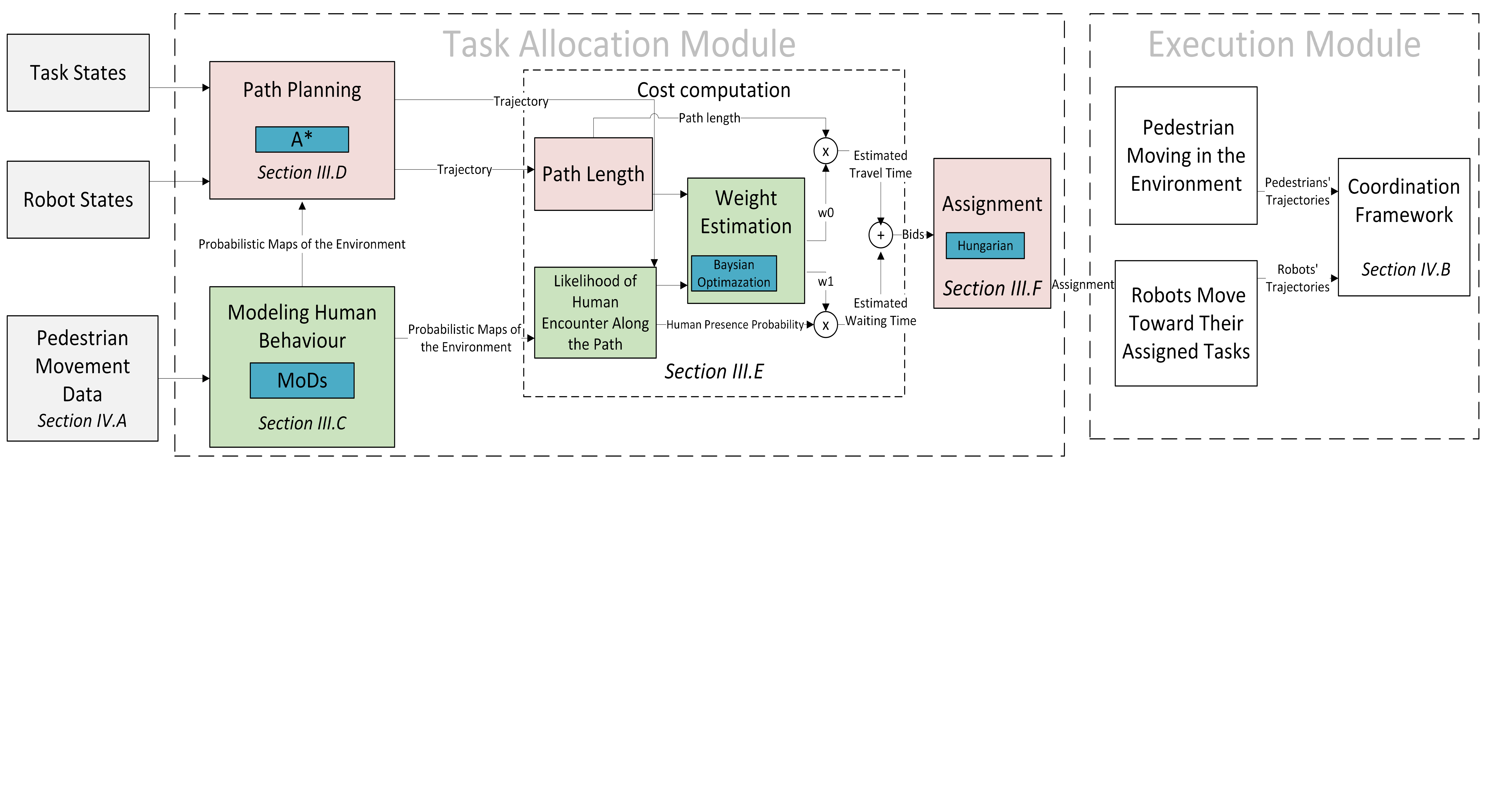}
    \caption{A framework that extends task allocation to human-aware task allocation. Red blocks are standard task allocation; green blocks are our contributions.}
    \label{framework}
    \vspace{-2em}
\end{figure*}
\section{Human Aware Task Allocation}

\label{sec:human aware task allocation}


\subsection {Problem Formulation}

In the MRTA problem in shared environments, each robot must navigate to its assigned task(s) while avoiding collisions with both static obstacles and moving pedestrians.
We consider a system comprising $n$ robots, denoted as $ R = \{r_1, r_2 ,..., r_n \} $, and equal number of tasks $ T = \{t_1 ,t_2 ,..., t_n \} $. Each robot $r_i$ can bid on task $t_j$.
The objective is to determine an efficient allocation that minimizes the maximum cost assigned to each robot, which is a proxy for minimizing the overall execution time. 
This allocation can be formulated as a binary integer programming problem:
\begin{align}
\begin{split}
Z =\min_{x_{ij}} \max_{i \in [1,n]} (c_{ij} x_{ij})\\
\text{s.t.~} \sum_{i=1}^n x_{ij} = 1 \quad \forall j \in [1,n]. \label{cost_function}
\end{split}
\end{align}
Each task $t_j$ is assigned to one robot, i.e.,
${\sum_{j=1}^n x_{ij} =1\quad\forall i \in [1,n]}$
, and each robot $r_i$ is assigned to one task
${x_{ij} \in \{0,1\} \quad \forall i,j \in [1,n].}$
$x_{ij}$ is a binary decision variable: 1 if robot $r_i$ is assigned to task $t_j$, 0 otherwise.
Finally, $c_{ij}$ is the cost for robot $r_i$ to task $t_j$.

\subsection {Architecture}
An overview of the system architecture is presented in \cref{framework}. The green boxes in the figure represent our contributions. Specifically, we propose to model human behavior through Maps of Dynamics (MoDs), (see Section III.C). From this, we derive the likelihood of human presence over time. Then we formulate a cost function(Section III.E), defined as a weighted sum of path length and likelihood of human encounters along the path. The weights for this cost function are estimated using Bayesian Optimization. Then, by leveraging an existing path planning algorithm (Section III.D), which takes the task and robot states as input and outputs trajectories, we estimate trajectory execution costs. Finally, these costs are submitted to an existing assignment algorithm (Section III.F) to allocate tasks to robots.

The system inputs are: robot and task states (their current coordination), and pedestrian movement data. Pedestrian movement data consists of historical records of timestamped pedestrian trajectories in the environment. The output of the system is an efficient allocation of tasks to robots, along with trajectories for each robot to execute its assigned tasks.

\deleted{An overview of the system architecture is presented in \cref{framework}. The system consists of four main components: modeling, path planning, cost computation, and task assignment. The environment modeling component uses historical human movement data to create probabilistic maps of dynamics in the environment. The path planning component takes the states of robots and tasks as input and computes trajectories from each robot to all tasks using the A* algorithm \cite{hart1968formal} on a static map, accounting for thresholded environmental constraints. These trajectories are then passed to the cost computation component, which evaluates a stochastic cost function as a weighted sum of path length and human encounter likelihood. The weights for these elements are optimized using Bayesian Optimization \cite{shahriari2015taking}. The computed costs are fed into the task assignment module, where a centralized dispatcher allocates tasks to robots efficiently based on the costs. Finally, the generated assignments are evaluated in a coordination module, where robots and humans coexist in the environment, accounting for actual travel and waiting times. Each component will be discussed in detail further.}
\vspace{-0.5em}
\subsection {Modeling human behavior}

\added{In dynamic environments, robots have limited information about the changes in their surroundings, such as the movement of humans. To address this, it is essential to model human movement patterns, enabling robots to anticipate and adapt to these changes effectively. Assuming that human motion exhibits consistent patterns, we can leverage historical data to model human movement behavior. To achieve this, the Modeling human behavior component, represented by the green block in the bottom left of \cref{framework}, utilizes MoDs to encode spatio-temporal patterns of human movements, enabling more informed task allocation.}

\deleted{As we have already pointed out, robots have limited information about the state of the environment. Thus we need to model dynamics in the environment using historical data, assuming that spatial-temporal patterns are periodical and will be repeated. The modeling human behavior component, which is represented with the green block at down left in \cref{framework}, address this challenge. we utilize MoDs, which store historical information about human motion. Thus, to anticipate the impact of human behaviors on robots' performance, we are utilizing MoDs, which store historical information about human motion.}
In this particular work, we utilize a time-dependent discrete probabilistic model, capturing the probability of encountering a person in a given cell. For each time interval $\Delta t$, we estimate the probability of encountering a person in cell $(i,j)$ by aggregating pedestrian data over time:
\begin{equation}
P(c_{i,j}, t + \Delta t) = \sum_{k \in \mathcal{K}_{i,j}} \frac{T_k} { \Delta t}.
\end{equation}
where $\mathcal{K}_{i,j}$ is the set of all pedestrian trajectories passing through cell $(i,j)$ during the interval $[t,t+\Delta t)$, and $T_k$ is the duration that pedestrian $k$ spends in cell$(i,j)$ during the interval.
\added{To ensure that overlapping time intervals are not double-counted, we only consider the unique presence of pedestrians in the cell. If multiple pedestrians are in the cell simultaneously, their overlapping time intervals are merged, and the total time is computed as the union of their individual intervals. This ensures that $P(c_{i,j})$ accurately reflects the likelihood of pedestrian presence without double counting.
The final output is a sequence of grid-based Maps of Dynamics (MoDs), which encodes the probability of encountering the pedestrian during the day.}
\vspace{-0.5em}
\subsection {Path Planning}
\added{The path planning component, shown by the red block in the top left of \cref{framework}, is responsible for generating feasible and safe trajectories for robots from their current states to all task states.}
\added{Since human presence introduces uncertainty into the environment, it is crucial to integrate information about human motion patterns into the planning process to improve both safety and efficiency. To achieve this goal, we incorporate MoDs into the path planning module. MoDs provide probabilistic maps that capture pedestrian movement patterns, enabling the system to make informed routing decisions. A key parameter in this process is the threshold parameter, $\delta$, which represents the maximum allowable probability of human presence in a given grid cell. If the probability in a cell exceeds this threshold, the planner proactively circumvents these high-risk areas. The choice of this threshold parameter plays a critical role in balancing efficiency and safety, and we discuss its selection in more detail later in section V.}
\added{To compute paths efficiently, we employ the A* algorithm \cite{hart1968formal}, which is well-known for its efficiency in finding optimal routes in grid-based environments.}

\deleted{We employ the A* algorithm for its efficiency in finding optimal routes in grid-based environments. To inform path planning of dynamics in the environment we incorporate MoDs into path planning, by introducing a threshold parameter, $\delta$. This threshold, $\delta$, represents the maximum allowable probability of encountering a human in a given grid cell. If the probability of human presence in a cell exceeds $\delta$, the planner adjusts the path to avoid high-risk areas.  This strategy is particularly advantageous in high-density areas, where rerouting to lower-risk zones may lengthen paths but significantly reduce potential waiting times; also, avoiding high-risk zones enhances safety.}
\vspace{-0.5em}
\subsection {Cost Computation}
\label{subsec: cost computation}
In this section, we elaborate on the cost computation module, represented with a dashed block in the middle of \cref{framework}. The cost computation combines the outputs of the previously described methods, the probabilistic maps from MoDs (Section III.C), and the trajectories generated by the path planning module (Section III.D). These inputs are used to compute the cost for each robot-task pair, which serves as the bid in the task allocation process.

Let us denote a matrix as $C = [c_{ij}]$ where $c_{ij}$ is the cost of robot $r_i$ performing task $t_j$. Each element $c_{ij}$ is the accumulated cost over the path starting from position $r_i$ and ending in position $t_j$. The cost of transitioning from one cell to the next is a weighted sum of the Euclidean distance and the likelihood of encountering humans, 
\begin{equation}
c_{ij} = \sum_{k} (w_0 \cdot d_k + w_1 \cdot \eta_k). \label{bid_cost}
\end{equation}
Here, $k$ represents segments id along the robot's path from its initial position to the task location, \(w_0\) scales the contribution of the Euclidean distance, and \(d_k\) is the Euclidean distance between the $(k-1)^{th}$ and $k^{th}$ cell in the robot's path.
Further, \(w_1\) scales the contribution of \(\eta_k\), which is a Bernoulli random variable equal to 1 if there is a human in a specific cell and 0 otherwise. Its expectation represents the average probability of encountering a human  $\mathbb{E}[\eta_k]\eqqcolon p_i$

The cost serves as each robot’s bid in the distributed system. The assignment component computes the assignment of tasks to robots, ensuring that the total cost is minimized.

To ensure that the cost accurately represents travel time in shared environments, we tune the cost function weights $w_0$ and $w_1$. However, evaluating different $w_0$ and $w_1$ configurations involves expensive simulations, making an exhaustive grid search impractical. To address this challenge, we employ Bayesian Optimization (BO) \cite{shahriari2015taking}, a well-suited tool for efficient global optimization of expensive-to-evaluate functions.  

In BO, the central idea is to model the objective function as a random function, placing a prior over it and updating it iteratively. We employ Gaussian Process Regression (GPR) for this purpose, as GPR provides a probabilistic framework that quantifies uncertainty in its predictions.

Gaussian Processes (GP) are non-parametric, probabilistic models that excel in capturing complex, non-linear relationships between variables~\cite{wang2023intuitive}.In the context of BO, GPR is used to model the relationship between the input parameters (e.g., weight configurations) and the observed outputs (e.g., mission time errors).  Considering a set of observed values denoted as $X$, an infinite number of potential functions can be fitted to these data points. The goal is to model this relationship using a latent function $f(x)$, which describes how the inputs $X$ relate to the observed outputs. GPR approaches this by considering all data points as observations from a multivariate Gaussian distribution, 
\begin{equation}
f(x) \sim {GP}(m(x), k(x, x')).
\label{gprgussian}
\vspace{-0.5em}
\end{equation}
In this framework, the GP is characterized by a mean function \( m(x) \) and a covariance function \( k(x, x') \), referred to as the kernel. \( m(x) \) provides the maximum likelihood estimate of the function at any point \( x \), and \( k(x, x') \) defines the correlation between function values at different points \( x \) and \( x' \), measuring how related \( x \) and \( x' \) are. Together, these functions allow GPR to capture both the shape and smoothness of the underlying function. In practice, we typically choose \( m(x) \equiv 0 \) without loss of generality. Thus, the GP model is a distribution over an infinite number of possible functions, with the specific shape and smoothness of these functions determined by the kernel function \( k(x, x') \). The choice of the kernel is an essential step in the smoothing process. Here, we adopt the Mat\'ern kernel, which is a common choice as it provides a balance between smoothness and capturing sudden changes~\cite{Kim2013gp}:
\begin{equation}
k(x, x') = \sigma^2 \frac{2^{1-\nu}}{\Gamma(\nu)} \left( \sqrt{2\nu} \frac{\|x - x'\|^2}{\ell} \right)^{\nu} K_{\nu} \left( \sqrt{2\nu} \frac{}{\ell} \right),
\vspace{-0.5em}
\end{equation}
where \( K_{\nu} \) is a modified Bessel function, \( \Gamma(\cdot) \) is the gamma function, \( \sigma_f^2 \) is the variance, controlling the variance of the function values, and \( \ell \) is the length-scale parameter, controlling how rapidly the correlation between points diminishes as they move farther apart in the input space.

Importantly, GPR refines its predictions by updating its posterior with new observations. This allows it to improve estimates while quantifying uncertainty. Instead of costly exhaustive simulations, GPR intelligently samples data, reducing trials while maintaining accuracy, enabling reliable weight estimation with fewer simulations.

In our context, the GP models the relationship between the weight parameters defined as $\mathbf{X} = \left\{ \mathbf{x}_i \in \mathbb{R}^2 \right\}_{i=1}^n$ and $\mathbf{y} = \left\{ \mathbf{y}_i \in \mathbb{R} \right\}_{i=1}^n$, which is the errors between estimated and simulated mission time. At each iteration, we update the GP model with new data points (weight configurations and their corresponding errors). The GP then provides a posterior distribution over possible weight configurations, allowing us to estimate both the expected performance and the uncertainty associated with different weight choices.

To select the next set of weights to evaluate, we employ an acquisition function that balances exploitation and exploration. Specifically, we use the Gaussian process upper confidence bound algorithm~\cite{srinivas2009gaussian} to determine the next weight configurations $\tilde{k}_i$ to be evaluated, 
\begin{equation}
\tilde{k}_i = \arg\max \mu(\tilde{k}_i) + \sqrt{\beta} \sigma(\tilde{k}_i),
\label{ucb}
\vspace{-0.5em}
\end{equation}
where $\mu(\tilde{k}_i)$ is the posterior mean for the weight configuration $\tilde{k}_i$, $\sigma(\tilde{k}_i)$ is the posterior variance for the weight configuration $\tilde{k}_i $, and $\beta $ is a parameter that controls the trade-off between exploration and exploitation.  We iteratively update the GP model and select new weight configurations until uncertainty falls below a threshold, enabling efficient exploration of the high-dimensional weight space.
\vspace{-0.5em}
\subsection {Assignment}

\added{The assignment process assigns tasks to robots to minimize the maximum cost for all robots, as formulated in \eqref{cost_function}. Each robot computes a bid locally for each task based on the stochastic cost function defined in \eqref{bid_cost}. The bid reflects the expected cost of traveling from the robot's current position to the task location. These bids are submitted to a centralized dispatcher, that coordinates the task allocation process.
The dispatcher collects bids and applies the Hungarian algorithm \cite{kuhn1955hungarian} to solve the assignment problem. The Hungarian algorithm is chosen for its efficiency in solving bipartite matching problems, making it suitable for real-time task allocation in dynamic environments.}

\section{Simulation Experiments Setup}
This section details our experiment setup for evaluating \acrshort{method name}, covering dynamic obstacle modeling, the coordination framework, and cost function weight estimation.


\subsection {Data and Maps of Dynamics Building}

\added{We evaluate our approach using the ATC dataset~\cite{brvsvcic2013person}, which provides real-world pedestrian motion data in a large and dynamic environment. Other available datasets, such as THÖR \cite{rudenko2020thor} and Edinburgh\cite{majecka2009statistical}, are not suitable for our study due to their limitations: THÖR provides only short-duration recordings, while the Edinburgh dataset is constrained to a small square environment with noisy measurements, making it unsuitable for evaluating scalability in realistic, large-scale settings.} 
\deleted{We use real pedestrian data from }The ATC dataset was recorded in a shopping mall in Osaka, Japan using a tracking system with multiple 3D range sensors covering an area over \SI{900}{\meter\squared}. The dataset was collected every Wednesday and Sunday from October 24, 2012, to November 29, 2013. It includes timestamps, person ID, position $(x, y, z)$, velocity, and angles of motion and facing. In this study, we use the timestamp, person ID, position $(x, y)$, velocity, and angle of motion.

To build MoDs, we utilize tracking data from Wednesday, November $7^{th}$, focusing on human trajectories within each 30-minute time window. The map was divided into grid cells with a resolution of \SI{0.05}{\meter}. Since the tracking data consists of single points representing individuals, we assumed each person occupies a larger area than a single point. Hence, a disk-shaped kernel of radius $r = 10$ cell around each tracked point was included in the probability update to account for the actual spatial area occupied by individuals. 

\subsection {Coordination Framework}
To evaluate the performance of our task allocation method comprehensively, we conducted experiments within a simulated environment using the multi-agent loosely-coupled coordination framework~\cite{pecora2018loosely}. In this environment, we replayed human trajectories and robot-to-goal trajectories, while the coordination framework handled intersections by enforcing precedence constraints. This approach ensures more accurate measurements of total mission time, waiting time, and failure rate, as it accounts for real-world interactions.

Specifically, the framework employs a centralized coordinator to manages robot movements along their assigned trajectories. By enforcing precedence constraints at intersections, the system prevents collisions and ensures safe navigation in critical sections. This capability allows the framework to handle both robot-robot and robot-pedestrian interactions effectively, making it a crucial component for evaluating task allocation in dynamic, human-populated environments. We modify the system to always give precedence to pedestrians to account for the worst-case scenario in the robots' waiting time. Also, there is no coordination between pedestrian-pedestrian intersections, to not disrupt the prerecorded data. 

We conducted experiments for fleets of size 5, 10, and 15 robots to assess the performance of different task allocation methods across varying fleet sizes. 
To evaluate the system's performance in the presence of humans, we replay human trajectories from December $12^{th}$ (a day different from those used for MoDs and weight estimation). For each scenario, we run 90 simulations, which are performed in three different time windows of the day.
In each experiment, we replay a different set of recorded human trajectories from the corresponding time interval to simulate realistic pedestrian movement. 
During these simulations, robots navigate assigned tasks while adapting to pedestrian dynamics, and we record their travel times and waiting times, capturing delays caused by human or robot interactions. These measurements provide a comprehensive evaluation of the system’s performance, allowing us to assess system’s performance under varying environmental conditions.
\vspace{-0.5em}
\subsection {Weight Estimation}
We now leverage the BO procedure from Section~\ref{subsec: cost computation} to iteratively tune the weights for task allocation. Initially, we begin with $w_0 = = w_1 = 1$ to assign robots to a set of tasks. With the robot trajectories in place, we conduct experiments in the coordination system to measure the actual time it takes for robots to reach their goals in the presence of humans. We consider the robots as car-like vehicles with a maximum velocity of \SI{1}{\meter\per\second}, maximum acceleration of \SI{1}{\meter\per\second\squared}, and a path resolution of \SI{0.05}{\meter}.
Human trajectories from November 28 (a different Wednesday than MoDs) were replayed in the coordination. 
We recorded travel time and waiting time for the robots in this setup to update the GP in the next iteration. Equation~\eqref{ucb} then proposes the next sampling candidate. Considering these weights, we rerun the allocation and coordination to record new data points for GPR. This iterative process continues until the uncertainty of the predicted weights becomes lower than 0.08. 

\begin{figure}[tb]
    \centering
    \includegraphics[scale=0.21]{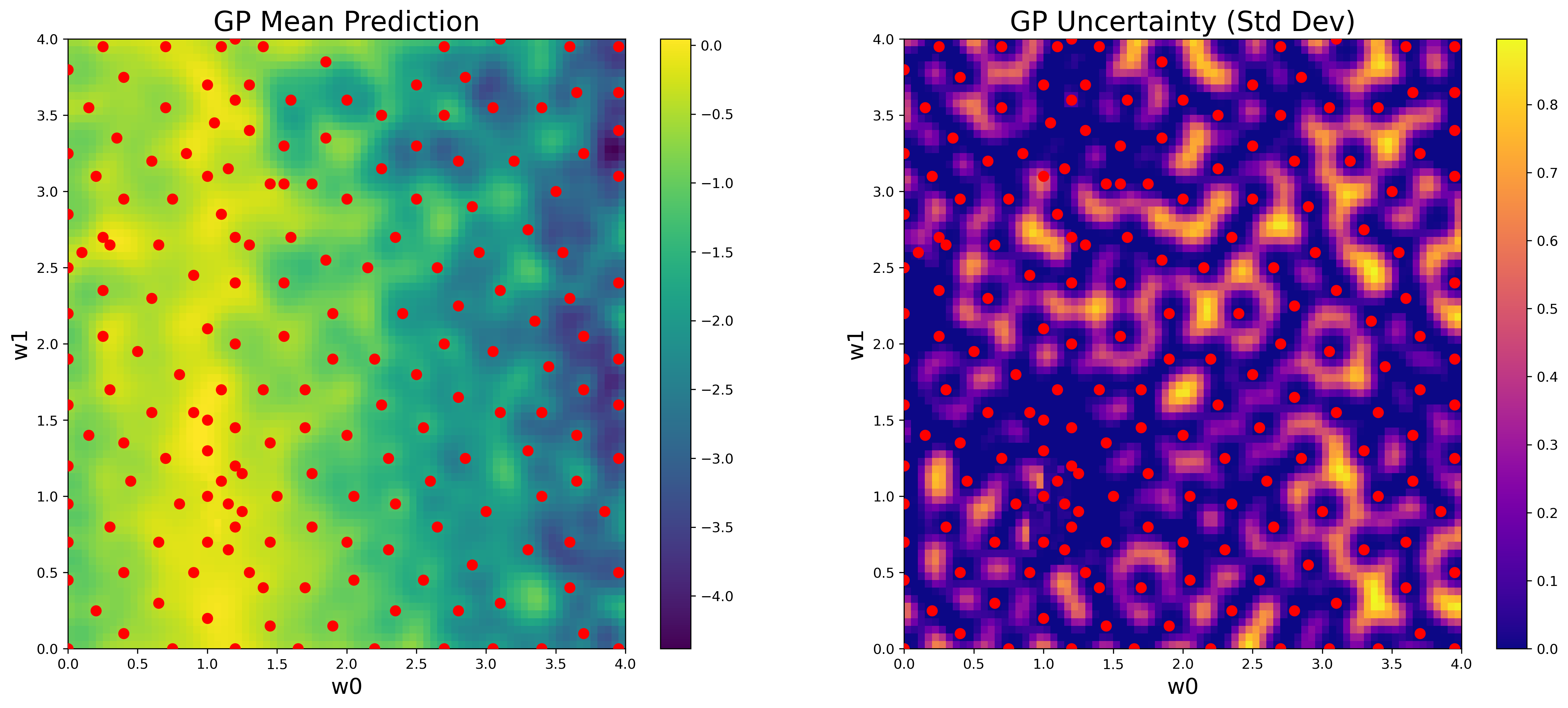}
    \caption{Visualisation of Gaussian Process Regression with $w_0$ and $w_1$ for $\beta = 150$ and $\ell = 0.08$, red dots mark locations of samples. The left figure shows the mean value of GP and the right figure shows the variance of GP.}
    \label{gp}
    \vspace{-2em}
    \end{figure}

To tune GPR hyperparameters, including the exploration-exploitation parameter $\beta$ and the length scale $\ell$, we relied on a heuristic search guided by the estimated weight uncertainty. After exploring different $\beta$ values, with a fixed length scale of $\ell = 0.1$, we chose $\beta = 150$. Subsequently, with fixed $\beta = 150$, we conducted experiments with different length scale values; we chose  $\ell = 0.08$.
Running BO with these parameters yielded the final weights of $w_0 =1.15$ and $w_1 = 0.95 $. \Cref{gp} shows the GPR output with these parameters.


\section{SIMULATION EXPERIMENTS AND RESULTS}

\begin{figure}[tb]
      \centering
      \includegraphics[trim=60 120 10 80, clip, scale=0.4]{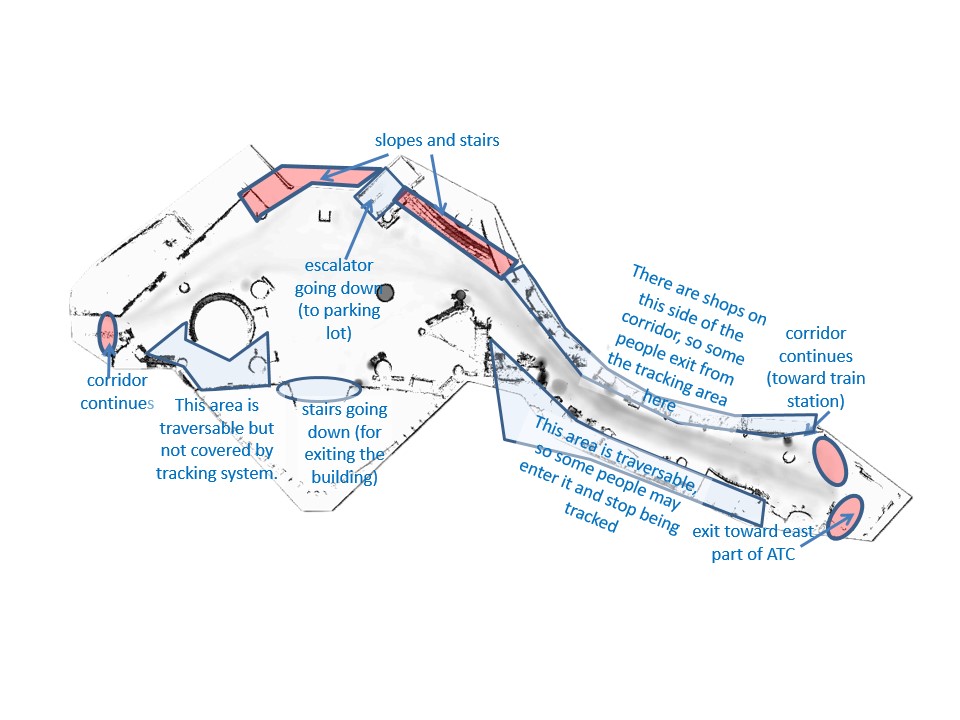}
      \caption{A top view of the ATC shopping mall, with static obstacles shown in black and semantic information indicating exits and detailed spatial features. The overlayed grayscale map shows Maps of Dynamics, where color intensity corresponds to the density of human presence, with darker tones indicating higher density, over a 30-minute interval on Wednesday, November 7, from 11:53:10 to 12:23:10}
      \label{mods}
      \vspace{-2em}
   \end{figure}

Our experiments evaluate the performance of \acrshort{method name} using real pedestrian data from the ATC dataset. \Cref{mods} shows a top view of the ATC shopping mall and MoDs generated by this data for a 30-minute time interval.

For evaluating \acrshort{method name}, we compare it to a standard auction-based allocation~\cite{lagoudakis2005auction}, which uses robots’ path lengths for bidding and is widely used in the field due to its simplicity, and efficiency. We also compare it with the HA-Alloc method introduced in~\cite{surma2021multiple}, which integrates human motion patterns into the allocation process. We conducted experiments with fleet sizes of 5, 10, and 15 robots, running \deleted{90} \added{1350} simulations in three different time windows of the day
T-0 (9:53:00 -- 10:23:00), T-1 (11:23:00 -- 11:53:00), and T-2 (15:53:00 -- 16:23:00), to account for varying human density. \added{At T-0, the human density was \(12 \pm 5\) pedestrians; at T-1 it was \(24 \pm 8\); and at T-2, it was \(21 \pm 6\).}
\vspace{-2pt}
\begin{table}[b]
\setlength{\abovecaptionskip}{2pt}
    \centering
    \caption{Comparison of Failure Rate for different methods}
    \label{tab:Failure_Rate}
    \begin{tabular}{@{}cc *{3}{c}}
    \toprule
  \multicolumn{1}{c}{}  &   &\multicolumn{3}{c}{Number of Robots}\\
    \multicolumn{1}{c}{} & &5    &10    &15      \\ \midrule
    \multirow{6}*{\rotatebox{90}{Method}}  
   & Path Based      &20\cellcolor{red!25}\%  & 37\cellcolor{red!25}\%    & 49\% \cellcolor{red!25}    \\ \cmidrule{2-5} 
   & HA-Alloc  &15\cellcolor{red!5}\%         & 33\cellcolor{red!15}\%    &   43\% \cellcolor{red!15}      \\ \cmidrule{2-5}
   & \acrshort{method name} $\delta=0.85$ &11\cellcolor{green!15}\%  & 26\cellcolor{green!5}\%     & 32\% \cellcolor{green!5}    \\ \cmidrule{2-5}
   & \acrshort{method name} $\delta=0.75$ &11\cellcolor{green!15}\%  & 23\cellcolor{green!25}\%     & 31\% \cellcolor{green!10}     \\ \cmidrule{2-5}
   & \acrshort{method name} $\delta=0.65$ &10\cellcolor{green!25}\%  & 24\cellcolor{green!10}\%    &  29\% \cellcolor{green!25}      \\ \cmidrule{2-5}
   & \acrshort{method name} $\delta=0.55$ &12\cellcolor{green!5}\%  & 22\cellcolor{green!25}\%     &  31\% \cellcolor{green!10}     \\ \bottomrule  
      \end{tabular}
      \vspace{-1em}
\end{table}
\raggedbottom

\setlength{\tabcolsep}{5pt} 
\renewcommand{\arraystretch}{1} 

\begin{table}[thpb]
    \centering
    \caption{Mission completion and waiting times for different methods (across fleet sizes)}
    \label{tab:Combined_Times_all}
    \footnotesize
    \begin{tabular}{@{}c c *{3}{c} | c *{3}{c}@{}}
    \toprule
    & & \multicolumn{3}{c}{Mission Completion Time} & \multicolumn{4}{c}{Waiting Time} \\
    \cmidrule(lr){3-5} \cmidrule(lr){6-9}
    & Method & 5 & 10 & 15 & & 5 & 10 & 15 \\ \midrule
    \multirow{6}*{\rotatebox{90}{Method}}  
    & Path Based         &158.7\cellcolor{red!25} & 174.4\cellcolor{red!25} & 299.6\cellcolor{red!25} & & 47.9\cellcolor{red!25} & 52.9\cellcolor{red!25} & 93.4\cellcolor{red!25} \\
    & HA-Alloc           &155.0\cellcolor{red!15} & 169.9\cellcolor{red!15} & 269.9\cellcolor{red!5}  & & 35.6\cellcolor{red!10} & 50.1\cellcolor{red!20} & 75.8\cellcolor{red!20} \\
    & \acrshort{method name} $\delta=0.85$ &140.5\cellcolor{green!5}  & 141.5\cellcolor{green!10} & 236.7\cellcolor{green!10} & & 32.6\cellcolor{green!5}  & 35.6\cellcolor{green!10} & 53.7\cellcolor{green!25} \\
    & \acrshort{method name} $\delta=0.75$ &130.4\cellcolor{green!20} & 137.2\cellcolor{green!15} & 255.0\cellcolor{green!5}  & & 22.2\cellcolor{green!25} & 31.0\cellcolor{green!20} & 60.4\cellcolor{green!15} \\
    & \acrshort{method name} $\delta=0.65$ &129.0\cellcolor{green!25} & 148.0\cellcolor{green!5}  & 234.1\cellcolor{green!25} & & 23.4\cellcolor{green!10} & 38.8\cellcolor{green!5}  & 53.7\cellcolor{green!25} \\
    & \acrshort{method name} $\delta=0.55$ &131.6\cellcolor{green!15} & 138.0\cellcolor{green!25} & 281.1\cellcolor{red!10}  & & 22.4\cellcolor{green!20} & 27.8\cellcolor{green!25} & 63.0\cellcolor{green!10} \\ \bottomrule
    \end{tabular}
\end{table}

Tables \ref{tab:Failure_Rate}, and \ref{tab:Combined_Times_all} present failure rates, mission times (the average total time taken by all robots to complete tasks), and waiting times (the average time spent by robots waiting for other robots or pedestrians) for the different methods. 
Failure rate refers to the percentage of tasks that were not successfully completed due to deadlock or timeout. The timeout was set to $10$ minutes. The 10-minute timeout exceeds the longest observed task time (under 4 minutes) to allow for unforeseen delays.

Table \ref{tab:Failure_Rate} shows HATA achieved the lowest failure rates across all fleet sizes, with improvements of up to 20\% compared to the path-based method and up to 14\% to HA-Alloc. Table \ref{tab:Combined_Times_all} shows that our method reduced overall mission completion times by up to 26\% compared to the path-based method and up to 19\%  to HA-Alloc. Our method also substantially reduced robot waiting times, as seen in Table \ref{tab:Combined_Times_all}. Improvements of up to 53\% over the path-based method and up to 41\% over HA-Alloc were observed.

 Meanwhile, mission times and waiting times vary with the threshold $\delta$, which represent the maximum allowable probability of encountering a human in a given cell. A smaller $\delta$ makes the system more conservative and circumvent risky cells which can lead to longer paths. Conversely, a higher $\delta$ allows the system to take more aggressive paths, optimizing mission times but increasing the risk of delays or failures due to human encounters. For instance, $\delta = 0.65$ yields a lower failure rate, meaning the system is more conservative and avoids risky situations where the probability of encountering a human exceeds 0.65. On the other hand, higher thresholds (e.g. $\delta = 0.75$) are more aggressive and can optimize mission times, indicating that careful selection of the threshold is crucial depending on the task requirements.

Table \ref{combined_times} compares mission and waiting times across intervals for a 5-robot fleet. Our method consistently outperforms others, showing robustness to varying human density. At busier intervals T-1 and T-2, thresholds 0.65 and 0.75 reduce mission times by up to 21\% and waiting times by up to 32\%, surpassing path-based and HA-Alloc methods. Similar trend in T-0 with slightly less pronounced improvements.

\begin{figure}[tb]
\setlength{\abovecaptionskip}{5pt}
    \centering
    \includegraphics[scale=0.38]{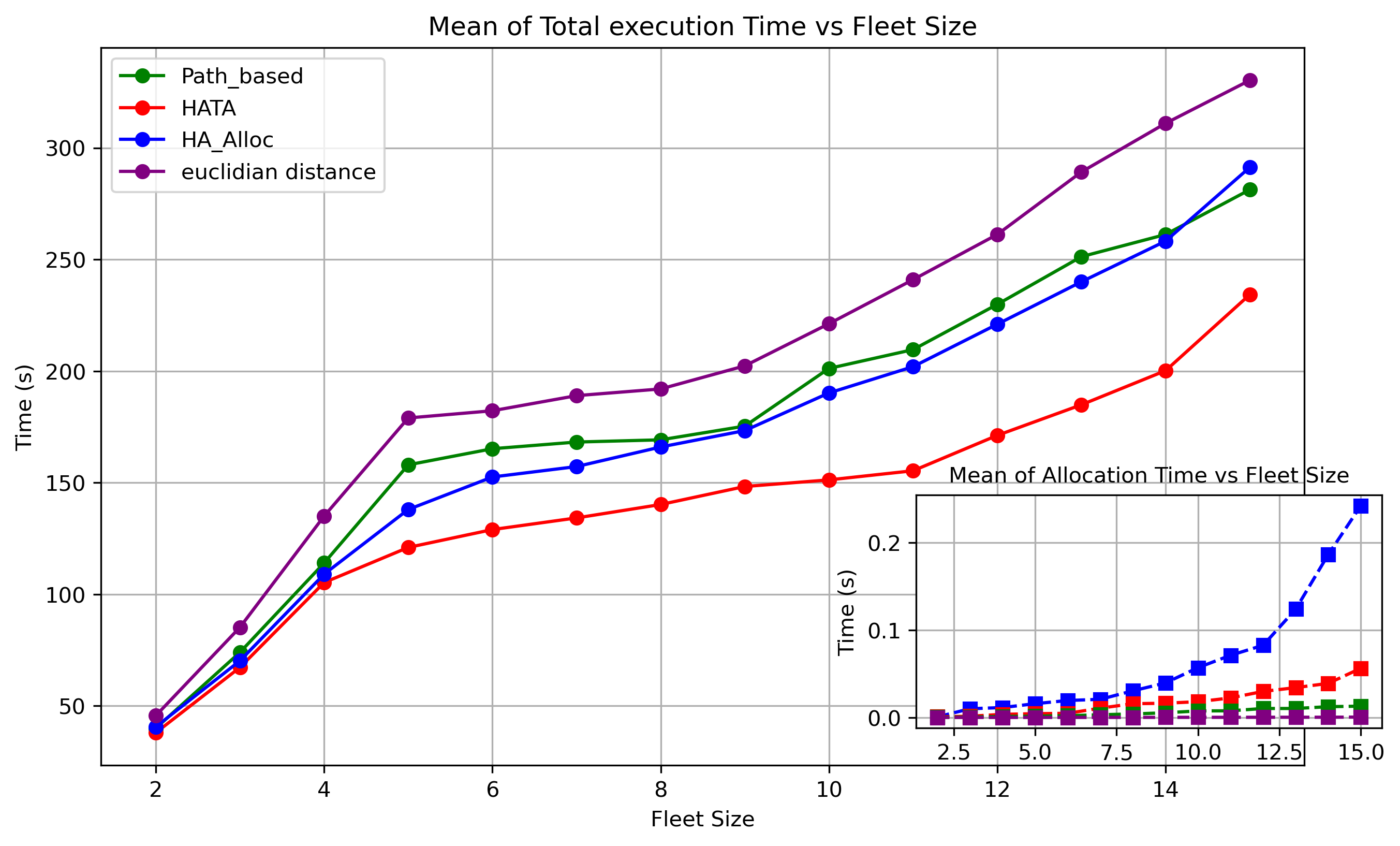}
    \caption{Comparison of total execution time and assignment time across different fleet sizes (2 to 15 robots) for four task allocation methods, HA-Alloc, HATA, Path-based, and Euclidean Distance. The main plot presents the total execution time averaged over 34 experiments, and the overlaid plot presents the assignment time, which includes bidding and task assignment. }
    \label{assignmenttime}
\end{figure}

While HATA improves the makespan, it has higher bid computation time than Euclidean or path-length methods due to human encounter computation along the trajectory using MoDs. However, compared to HA-Alloc, HATA demonstrates a clear advantage in computational efficiency and scalability.
More importantly, the additional bid computation time is minimal relative to the total execution time and does not create a bottleneck.To illustrate, the main plot in \cref{assignmenttime} shows average total execution time over 34 experiments, while the overlaid plot compares average assignment times of HATA, HA-Alloc, Path-based, and Euclidean Distance for varying fleet sizes. 
The assignment time, which includes bidding and task assignment but excludes path planning (as it depends on the specific path-planning approach), remains significantly lower than the total execution time. This reinforces that the added computational cost of bidding is negligible compared to the overall benefits of improved task allocation efficiency.

\begin{table}[tb]
\centering
\caption{Mission and waiting times for different methods across time intervals with varying human density for a fleet of 5 robots.}
\label{combined_times}
\footnotesize
\adjustbox{max width=\linewidth}{
\begin{tabular}{@{}lcccccc@{}}
\toprule
\multirow{2}{*}{Method} & \multicolumn{3}{c}{Mission Time} & \multicolumn{3}{c}{Waiting Time} \\ 
\cmidrule(lr){2-4} \cmidrule(lr){5-7}
 & T\_0 & T\_1 & T\_2 & T\_0 & T\_1 & T\_2 \\
\midrule
Path Based & 115.38 & 192.95 & 173.36 & 23.11 & 56.71 & 92.72 \\
HA-Alloc & 123.25 & 178.86 & 164.79 & 33.04 & 38.24 & 98.88 \\
\acrshort{method name} $\delta=0.85$ & 104.03 & 165.64 & 155.53 & 16.76 & 33.40 & 77.23 \\
\acrshort{method name} $\delta=0.75$ & 112.63 & 133.36 & 149.70 & 13.50 & 22.46 & 63.36 \\
\acrshort{method name} $\delta=0.65$ & 97.50 & 152.19 & 140.93 & 15.94 & 35.71 & 84.54 \\
\acrshort{method name} $\delta=0.55$ & 108.45 & 137.62 & 154.86 & 15.28 & 30.71 & 57.40 \\
\bottomrule
\end{tabular}
}
\vspace{-2.5em}
\end{table}

Our experiments were limited to fleets of up to 15 robots due to coordination framework constraints, highlighting the need for more scalable solutions. As Figure 4 shows, HATA maintains low planning times and scales well, but higher robot densities can cause congestion. Handling this requires integrated coordination during task allocation, which is beyond this paper’s scope.

\section{CONCLUSION AND FUTURE WORKS}

This work aimed to improve Multi-Robot Task Allocation (MRTA) in human-populated environments by incorporating human motion patterns into the decision-making process. We proposed \acrshort{method name}, which leverages Maps of Dynamics (MoDs) to model motion patterns and integrate this knowledge into a stochastic cost function. This enhancement allows robots to anticipate and avoid high-density areas, reducing waiting times and improving overall mission efficiency. Experiments' results confirm that this approach outperforms both path-based and HA-Alloc methods in dynamic environments, particularly when human-robot interactions are frequent.

\deleted{\acrshort{method name} addresses key limitations in traditional MRTA methods by incorporating human motion patterns using MoDs into the decision-making process. This enhancement allows robots to anticipate and avoid high-density areas, reducing waiting times and improving overall mission efficiency. The results of extensive experiments confirm that this approach outperforms both path-based and HA-Alloc methods in dynamic environments, particularly when human-robot interactions are frequent.}

\added{While results indicate that \acrshort{method name} scales well, our evaluation was limited to 15 robots due to constraints in the coordination framework, which can handle up to 20 agents (including both robots and pedestrians) and no alternative was available that could efficiently manage robot-human interactions at a larger scale. Future works could explore more scalable coordination mechanisms. Additionally, while our method effectively incorporates human dynamics, an interesting direction for future works is to investigate whether a more comprehensive approach to modeling human behavior would further enhance efficiency.}

\deleted{Our results demonstrate that our framework significantly enhances the efficiency of multi-robot systems in dynamic, human-populated environments. It paves the way for incorporating richer semantic information, such as human movement direction, which could further refine the system's ability to anticipate and react to human behavior, making the task allocation even more effective.}



\bibliographystyle{IEEEtran}
\bibliography{HATA_ref2}


\thispagestyle{fancy}	
\pagestyle{empty}
\end{document}